\documentclass[11pt,a4paper]{article}
\usepackage[hyperref]{conll-2019}
\usepackage{times}
\usepackage{latexsym}
\usepackage{graphicx}      
\usepackage{url}
\usepackage{algorithm, algpseudocode}
\usepackage{varwidth}
\usepackage{amsfonts, amsmath} 
\usepackage{multicol}
\usepackage{multirow}
\usepackage{subfigure}
\usepackage{enumitem}
\setlist[enumerate]{itemsep=0mm}

\algrenewcommand\textproc{}
\makeatletter
\algrenewcommand\ALG@beginalgorithmic{\footnotesize}
\algrenewcommand\algorithmiccomment[2][\normalsize]{{#1\hfill\(\triangleright\) #2}}
\makeatother

\aclfinalcopy 

\newcommand\Wapprox{\texttt{SampleApprox}\,}
\newcommand\Uapprox{\texttt{CapSampleApprox}\,}
\newcommand\NNFST{NN2WFA\,}
\newcommand{\RR}{\mathbb{R}}
\newcommand{\indic}{1}
\newcommand{\tcO}{\tilde{\mathcal{O}}}
\title{Federated Learning of N-gram Language Models}

\author{Mingqing Chen, Ananda Theertha Suresh, Rajiv Mathews, Adeline Wong,\\
  {\bf Cyril Allauzen, Fran{\c{c}}oise Beaufays, Michael Riley}\\
  Google, Inc.\\
  {\tt \{mingqing,theertha,mathews,adelinew,allauzen,fsb,riley\}} \\
  \tt {@google.com}}

\date{}

\begin{document}
\maketitle
\begin{abstract}
  We propose algorithms to train production-quality n-gram
  language models using federated learning. Federated learning is a distributed
  computation platform that can be used to train global models for portable
  devices such as smart phones. Federated learning is especially relevant for
  applications handling privacy-sensitive data, such as virtual keyboards,
  because training is performed without the users' data ever leaving their
  devices. While the principles of federated learning are fairly generic, its
  methodology assumes that the underlying models are neural networks. However,
  virtual keyboards are typically powered by n-gram language models for latency
  reasons.

  We propose to train a recurrent neural network language model using
  the decentralized \texttt{FederatedAveraging} algorithm and to
  approximate this federated model server-side with an n-gram model
  that can be deployed to devices for fast inference. Our technical
  contributions include ways of handling large vocabularies,
  algorithms to correct capitalization errors in user data, and
  efficient finite state transducer algorithms to convert word
  language models to word-piece language models and vice versa. The
  n-gram language models trained with federated learning are compared
  to n-grams trained with traditional server-based algorithms using
  A/B tests on tens of millions of users of a virtual keyboard.
  Results are presented for two languages, American English and
  Brazilian Portuguese. This work demonstrates that high-quality
  n-gram language models can be trained directly on client mobile
  devices without sensitive training data ever leaving the devices.
\end{abstract}
\section{Introduction}
\subsection{Virtual keyboard applications}
\begin{figure}
  \centering
  \includegraphics[width=0.85\columnwidth]{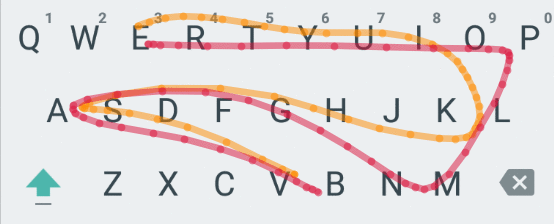}
  \caption{Glide trails are shown for two spatially-similar words:
    ``Vampire'' (in red) and ``Value'' (in orange). Viable decoding candidates
    are proposed based on context and language model scores.}
  \label{fig:glide_trail_similar_words}
\end{figure}

Virtual keyboards for mobile devices provide a host of functionalities from
decoding noisy spatial signals from tap and glide typing inputs to providing
auto-corrections, word completions, and next-word predictions. These features
must fit within tight RAM and CPU budgets, and operate under strict latency
constraints. A key press should result in visible feedback within about 20
milliseconds~\citep{Ouyang17Mobile, nsm}. Weighted finite-state transducers
have been used successfully to decode keyboard spatial signals using a
combination of spatial and language
models~\citep{Ouyang17Mobile, Hellsten17Transliterated}.
Figure~\ref{fig:glide_trail_similar_words}
shows the glide trails of two spatially-similar words. Because of the similarity
of the two trails, the decoder must rely on the language model to discriminate
between viable candidates.
For memory and latency reasons, especially on low-end devices, the language
models are typically based on n-grams and do not exceed ten megabytes. A
language model (LM) is a probabilistic model on words. Given previous words
$x_1,x_2,\ldots, x_{m-1}$, an LM assigns a probability to the new words, i.e.
$p(x_{m} | x_{m-1},\ldots, x_{1})$. An n-gram LM is a Markovian distribution of
order $n-1$, defined by
\[
p(x_{m} | x_{m-1},\ldots,
x_{1}) = p(x_{m} | x_{m-1},\ldots, x_{m-n+1}),
\]
where $n$ is the order of the n-gram. For computation and memory efficiency,
keyboard LMs typically have higher-order n-grams over a subset of the
vocabulary, e.g. the most frequent $64$K words, and the rest of the vocabulary
only has unigrams. We consider n-gram LMs that do not exceed
$1.5$M n-grams and include fewer than $200$K unigrams.

N-gram models are traditionally trained by applying a smoothing method to n-gram
counts from a training corpus~\cite{chen1999empirical}. The highest quality
n-gram models are trained over data that are well-matched to the desired
output~\cite{intelligent_training_data}. For virtual keyboards, training over
users' typed text would lead to the best results. Of course, such data are very
personal and need to be handled with care.

\subsection{Federated learning}

\begin{figure}
  \centering
  \includegraphics[width=0.99\columnwidth]{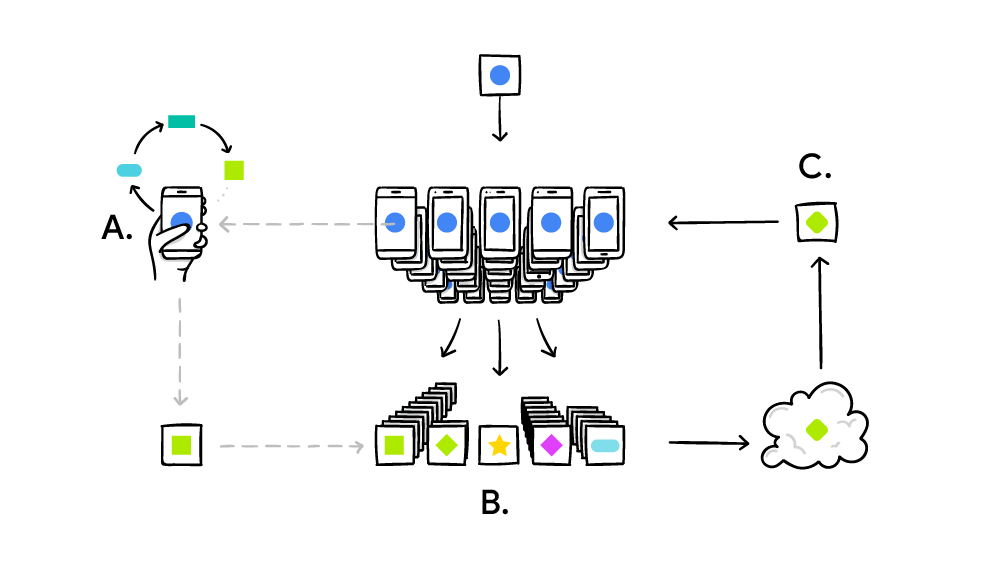}
  \caption{An illustration of the federated learning process from
           \citet{fedblog}: (A) client devices compute SGD updates on
           locally-stored data, (B) a server aggregates the client updates
           to build a new global model, (C) the new model is sent back to
           clients, and the process is repeated.}
  \label{fig:illustration_fed}
\end{figure}

We propose to leverage \emph{Federated Learning}
(FL)~\cite{konevcny2016federated, konevcny2016federatedb}, a technique where
machine learning models are trained in a decentralized manner on end-users'
devices, so that raw data never leaves these devices. Only
targeted and ephemeral parameter updates are aggregated on a
centralized server. Figure~\ref{fig:illustration_fed} provides an illustration
of the process.
Federated learning for keyboard input was previously explored in
\citet{hard2018federated}, in which a federated recurrent neural network (RNN)
was trained for next-word prediction. However, latency constraints prevent the
direct use of an RNN for decoding. To overcome this problem, we propose to
derive an n-gram LM from a federated RNN LM model and use that n-gram LM for
decoding. Specifically, the approximation algorithm is based on \Wapprox, which
was recently proposed in \citet{suresh2018approx, suresh2019distilling}. The proposed approach has
several advantages:

\noindent\textbf{Improved model quality:} Since the RNN LM is trained directly
on domain-matched user data, its predictions are more likely to match
actual user behavior. In addition, as shown in \citet{suresh2018approx},
an n-gram LM approximated from such an RNN LM is of higher quality than an n-gram
LM trained on user data directly.

\noindent\textbf{Minimum information transmission:} In FL, only the minimal
information necessary for model training (the model parameter deltas) is
transmitted to centralized servers. The model updates contain much less
information than the complete training data.

\noindent\textbf{Additional privacy-preserving techniques:}
FL can be further combined with privacy-preserving techniques such as secure
multi-party computation~\cite{Bonawitz2017} and differential
privacy~\cite{McMahan2017LearningDP, agarwal2018cpsgd, abadi2016deep}. By the
post-processing theorem, if we train a single differentially private recurrent
model and use it to approximate n-gram models, all the distilled models will
also be differentially private with the same
parameters~\cite{dwork2014algorithmic}.

For the above reasons, we have not proposed to learn n-gram models directly
using \texttt{FederatedAveraging} of n-gram counts for all orders.

\section{Outline}
\label{sec:outline}
The paper is organized along the lines of challenges associated with converting
RNN LMs to n-gram LMs for virtual keyboards: the feasibility of training neural
models with a large vocabulary, inconsistent capitalization in the training
data, and data sparsity in morphologically rich languages. We elaborate on each
of these challenges below.

\noindent\textbf{Large vocabulary:}
Keyboard n-gram models are typically based on a carefully hand-curated
vocabulary to eliminate misspellings, erroneous capitalizations, and other
artifacts. The vocabulary size often numbers in the hundreds of thousands.
However, training a neural model directly over the vocabulary is memory
intensive as the embedding and softmax layers require space
$\left | \mathcal{V} \right | \times N_e$, where $\left | \mathcal{V} \right |$
is the vocabulary size and $N_e$ is the embedding dimension.
We propose a way to handle large vocabularies for federated models in Section~\ref{sec:unigram}.

\noindent\textbf{Incorrect capitalization:} In virtual keyboards, users often
type with incorrect casing (e.g. ``She lives in new york'' instead of
``She lives in New York"). It would be desirable to decode with the correct
capitalization even though the user-typed data may be incorrect. Before
the discussion of capitalization, the \Wapprox algorithm is reviewed in
Section~\ref{sec:wapprox}. We then modify \Wapprox to infer capitalization
 in Section~\ref{sec:capitalization}.

\noindent\textbf{Language morphology:}
Many words are composed of root words and various morpheme components, e.g.
``crazy'', ``crazily'', and ``craziness''. These linguistic features are
prominent in morphologically rich languages such as Russian. The presence of a
large number of morphological variants increases the vocabulary size and data
sparsity ultimately making it more difficult to train neural models.
Algorithms to convert between word and word-piece models are discussed in
Section~\ref{sec:word_piece}.

Finally, we compare the performance of word and word-piece models and present
the results of A/B experiments on real users of a virtual keyboard in
Section~\ref{sec:experiments}.

\section{Unigram distributions}
\label{sec:unigram}
Among the $200$K words in the vocabulary, our virtual keyboard models
only use the top $64$K words in the higher-order n-grams. We train the neural
models only on these most frequent words and train a separate unigram model over
the entire vocabulary. We interpolate the two resulting models to obtain the
final model for decoding.
\subsection{Collection}
Unigrams are collected via a modified version of the \texttt{FederatedAveraging}
algorithm. No models are sent to client devices. Instead of returning gradients
to the server, counting statistics are compiled on each device and returned. In
our experiments, we aggregate over groups of approximately 500 devices per
training round. We count a unigram distribution $U$ from a 
whitelist vocabulary by $U = \sum_i w_i C_i$, where $i$ is the index over
devices, $C_i$ are the raw unigram counts collected from a single device $i$,
and $w_i$ is a weight applied to device $i$.

To prevent users with large amounts of data from dominating the unigram
distribution, we apply a form of L1-clipping:
\begin{equation}
  w_i = \frac{\lambda}{\max(\lambda, \sum C_i)},
\end{equation}
where $\lambda$ is a threshold that caps each device's contribution. When
$\lambda = 1$, L1-clipping is equivalent to equal weighting. The limit
$\lambda \to \infty$ is equivalent to collecting the true counts, since
$w_i \to 1$.
\subsection{Convergence}
Convergence of the unigram distribution is measured using the unbiased
chi-squared statistic (for simplicity, referred to as the $Z$-statistic) defined
in \citet{bhattacharya2015similarity}, the number of unique unigrams seen, and a
moving average of the number of rounds needed to observe new unigrams.

\begin{figure}
\centering
\subfigure[]{
\includegraphics[width=.46\linewidth]{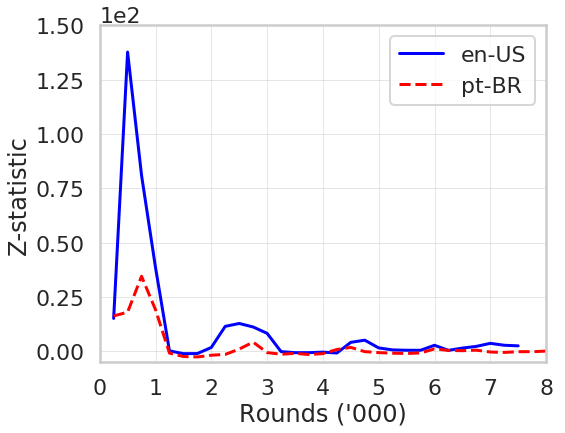}
\label{fig:distro_convergence}}
\subfigure[]{
\includegraphics[width=.47\linewidth]{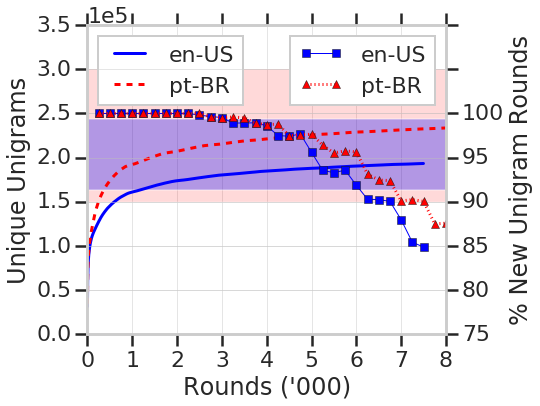}
\label{fig:learning_tail_words}}
\caption{Unigram distribution convergence. Note that by 3000 rounds, the unigram
distribution is stable, but the model is still learning new tail unigrams.}
\end{figure}

Figure~\ref{fig:distro_convergence} shows the overall distributional convergence
based on the $Z$-statistic. At round $k$, unigram counts after $k/2$ and $k$
rounds are compared.
Figure~\ref{fig:learning_tail_words} plots the number of whitelist vocabulary
words seen and a moving average of the number of rounds containing new unigrams.
New unigrams are determined by comparing a round $k$ with all rounds
through $k-1$ and noting if any new words are seen.
The shaded bands range from the LM's unigram capacity to the size of
the whitelist vocabulary.

\subsection{Experiments}
Since the whitelist vocabulary is uncased, capitalization normalization is applied
based on an approach similar to Section~\ref{sec:capitalization}.
We then replace
the unigram part of an n-gram model with this distribution to produce the
final LM.

In A/B experiments, unigram models with different L1-clipping thresholds are
compared against a baseline unigram model gathered from centralized log data.
Results are presented in Table~\ref{table:expts_l1clipping}. Accuracy is
unchanged and OOV rate is improved at $\lambda = 1$ and $\lambda = 1K$.
\begin{table}[]
\centering
\begin{tabular}{c|cccc}
Model             & acc@1 [\%]          & OOV rate [\%]       \\ \hline
baseline          & 8.14                & 18.08               \\ \hline
$\lambda = 1$     & $+0.19 \pm 0.21$    & $-1.33 \pm 0.75$    \\
$\lambda = 1K$    & $+0.11 \pm 0.24$    & $-1.06 \pm 0.66$    \\
$\lambda = 5K$    & $-0.08 \pm 0.26$    & $-0.78 \pm 0.93$    \\
\end{tabular}
\caption{Relative change with L1-clipped unigrams on live traffic of en\_US
         users on the virtual keyboard. Quoted 95\% confidence
         intervals are derived using the jackknife method with user buckets.}
\label{table:expts_l1clipping}
\end{table}

Before we discuss methods to address inconsistent capitalization and data sparsity
in morphologically rich languages, we review \Wapprox.
\section{Review of \Wapprox}
\label{sec:wapprox}
\Wapprox, proposed in \citet{suresh2018approx, suresh2019distilling}, can be used to approximate a
RNN as a weighted finite automaton such as an n-gram model. A {\em
weighted finite automaton} (WFA) $A = (\Sigma, Q, E, i, F)$ over
$\RR_{+}$ (probabilities) is given by a finite alphabet $\Sigma$
(vocabulary words), a finite set of states $Q$ (n-gram contexts), an
initial state $i \in Q$ (sentence start state), a set of final states
$F \in Q$ (sentence end states), and a set of labeled transitions $E$
and associated weights that represent the conditional probability of
labels (from $\Sigma$) given the state (list of n-grams and their
probabilities). WFA models allow a special
\emph{backoff} label $\varphi$ for succinct representation as follows.
Let $L[q]$ be the set of labels on transitions from state $q$. For $x \in L[q]$,
let $w_q[x]$, be the weight of the transition of $x$ at state $q$ and $d_q[x]$
be the destination state. For a label $x$ and a state $q$,

\begin{align*}
p(x | q) & = w_q[x] &\text{ if }x \in L[q], \\ & =w_{q}[\varphi] \cdot
p(x | d_q[\varphi]) & \text{ otherwise}.
\end{align*}
In other words, $\varphi$ is followed if $x \notin L[q]$.
The definition above is consistent with that of backoff n-gram
models~\cite{chen1999empirical}. Let $B(q)$ denote the set of states from which
$q$ can be reached by a path of backoff labels and let $q[x]$ be the first state
at which label $x$ can be read by following a backoff path from $q$.

Given an unweighted finite automaton $A$ and a neural model, \Wapprox finds the
probability model on $A$ that minimizes the Kullback-Leibler (KL) divergence
between the neural model and the WFA. The algorithm has two steps: a counting
step and a KL minimization step. For the counting step, let
$\bar{x}(1), \bar{x}(2), \ldots, \bar{x}(k)$ be $k$ independent samples
from the neural model. For a sequence $\bar{x}$, let $x_i$ denote the
${i}^{\text{th}}$ label and $\bar{x}^{i} = x_1, x_2,\ldots, x_i$ denote the
first $i$ labels. For every $q \in Q$ and $x \in \Sigma$, the algorithm computes
$C(x, q)$ given by
\[
\sum_{q' \in B(q)} \sum^m_{j=1} \sum_{i \geq 0}
\indic_{q(\bar{x}^i(j)) = q', q = q'[x]} \cdot
p_\text{nn}(x | \bar{x}^{i}(j)).
\]
We illustrate this counting with an example. Suppose we are interested in the
count of the bi-gram \textrm{New York}. Given a bi-gram LM, \Wapprox
generates $m$ sentences and computes
\[
C(\textrm{York}, \textrm{New})= \sum_{j, i :x_i(j)= \textrm{New}}
p_{\text{nn}}(\textrm{York} | \bar{x}^i(j)).
\]
In other words, it finds all sentences that have the
word \textrm{New}, observes how frequently \textrm{York} appears
subsequently, and computes the conditional probability. After
counting, it uses a difference of convex (DC) programming based
algorithm to find the KL minimum solution. If $\ell$ is the average
number of words per sentence, the computational complexity of counting
is $\tcO(k \cdot \ell \cdot |\Sigma|)$~\footnote{$a_n = \tcO(b_n)$,
means $a_n \leq b_n \cdot \text{poly}\log(n), \forall n \geq n_0$.}
and the computational complexity of the KL minimization is $\tcO(|E| +
|Q|)$ per iteration of DC programming.

\section{Capitalization}
\label{sec:capitalization}
As mentioned in Section~\ref{sec:outline}, users often type with incorrect
capitalization. One way of handling incorrect capitalization is to store an
on-device capitalization normalizer \cite{beaufays2013language} to correctly
capitalize sentences before using them to train the neural model. However,
capitalization normalizers have large memory footprints and are not suitable for
on-device applications. To overcome this, the neural model is first trained on uncased
user data. \Wapprox is then modified to approximate cased n-gram models from
uncased neural models.

As before, let $\bar{x}(1), \bar{x}(2), \ldots, \bar{x}(k)$ be $k$ independent
(uncased) samples from the neural model. We capitalize them correctly at the
server using~\citet{beaufays2013language}. Let $\bar{y}(1), \bar{y}(2),\ldots
\bar{y}(k)$ represent the corresponding $k$ correctly capitalized samples. Let
$p_\text{cap}$ be another probability model on non-user data that approximates
the ratio of uncased to cased probabilities given a context. Given a label $y$,
let $u(y)$ be the uncased symbol. For example, if $y$ is \textrm{York}, then
$u(y)$ is \textrm{york}. With the above definitions, we modify the counting step
of \Wapprox as follows:
\[
\sum_{q' \in B(q)} \sum^m_{j=1} \sum_{i \geq 0}
\indic_{q(\bar{y}^i(j)) = q', q = q'[y]} \cdot
\tilde{p}(y | \bar{y}^{i}(j)),
\]
where $\tilde{p}(y | \bar{y}^{i}(j))$ is given by
\[
p_\text{nn}(u(y) | u(\bar{y}^{i}(j))) \cdot
\frac{p_\text{cap}(y| \bar{y}^{i}(j) )}{\sum_{y' : u(y') = u(y)}
  p_\text{cap}(y' | \bar{y}^{i}(j))}.
\]
We refer to this modified algorithm as \Uapprox. We note that
word-piece to word approximation incurs an additional computation cost
of $\tcO((|E| + |Q| + |\Delta|)\ell)$, where $\Delta$ is the number of
words, $E$ and $Q$ are the set of arcs and set of states in the word
n-gram model, and $\ell$ is the maximum number of word-pieces per word.

\section{Morphologically rich languages}
\label{sec:word_piece}
To train neural models on morphologically rich languages, subword segments such
as byte-pair encodings or
word-pieces~\cite{shibata1999byte,schuster2012japanese, taku2018} are typically
used. This approach assigns conditional probabilities to subword segments,
conditioned on prior subword segments. It has proved successful
in the context of speech recognition~\cite{chiu2018state} and
machine translation~\cite{wu2016google}. Following these successes, we propose
to train RNN LMs with word-pieces for morphologically rich languages.

\begin{figure}[t]
{\footnotesize \begin{center}
\begin{tabular}{c@{\hspace{0ex}}c}
\scalebox{.45}{\includegraphics{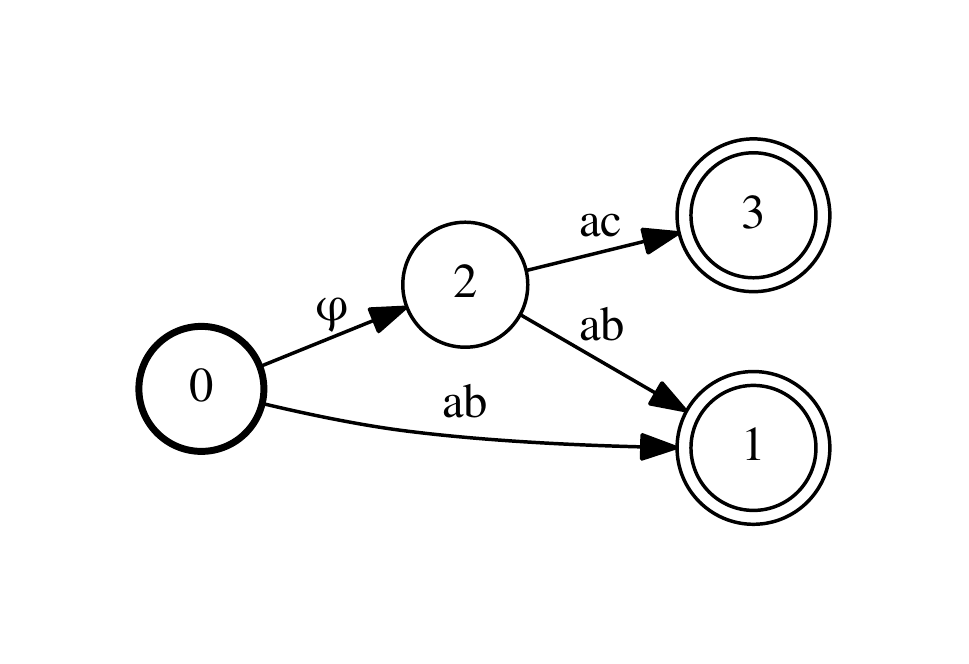}} &
\raisebox{1ex}{\scalebox{.45}{\includegraphics{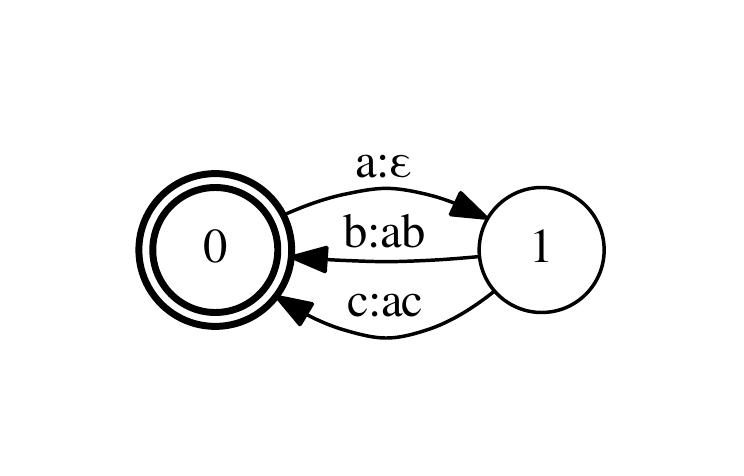}}} \\[-5ex]
(a) & (b) \\[-2ex]
\multicolumn{2}{c}{\scalebox{.45}{\includegraphics{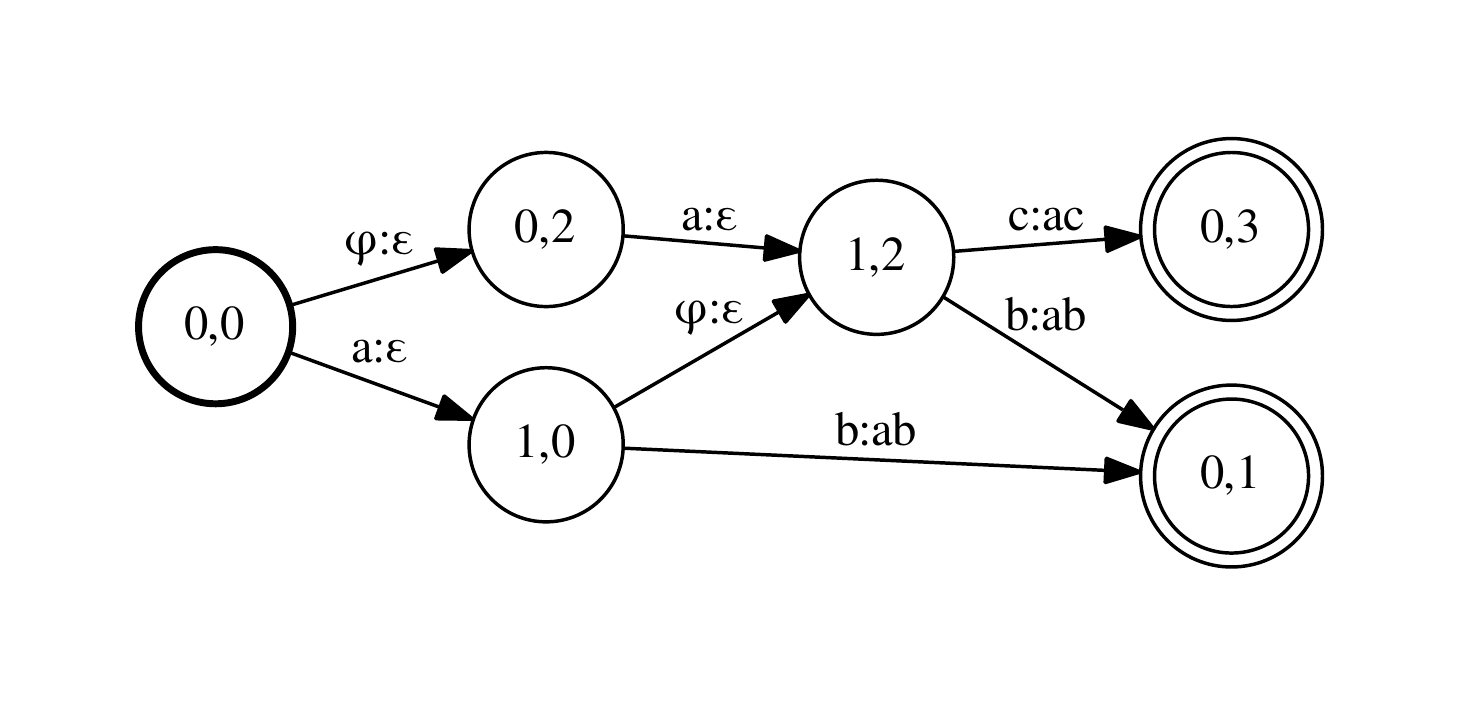}}} \\[-5ex]
\multicolumn{2}{c}{(c)}
\end{tabular}
\end{center}}
\caption{\label{fig:wordpiece}The (a) WFA $A$ and WFSTs (b) $T$ and (c) $B$ for
the word vocabulary $\{ab, ac\}$ and word-piece vocabulary $\{a, b, c\}$.
Initial states are represented by bold circles and final states by double
circles.}
\end{figure}

We apply the word-piece approach of \citet{taku2018}, which computes a
word-piece unigram LM using a word-piece inventory $\mathcal{V_P}$. Each
word-piece $x_i\in\mathcal{V_P}$ is associated with a unigram probability
$p(x_i)$.
For a given word $y$ and its possible segmentation candidates,
the word is encoded with the segmentation that assigns the highest probability.

Throughout this paper we apply $4$K, $16$K, and $30$K as the word-piece
inventory sizes. These values lie within a range that provides good
trade-off between the LSTM embedding size and the richness of the language
morphology. We apply $100\%$ character coverage to include all the
symbols that appeared in the unigram distribution
(Section~\ref{sec:unigram}), including the common English letters,
accented letters e.g. \'e, \^{o}, and digits. Accented letters are important for
languages like Portuguese. For fast decoding, the n-gram models still need to be
at the word-level, since word-piece n-gram models increase the depth of
the beam-search during decoding. We convert the word n-gram topology to an
equivalent word-piece WFA topology and use \Wapprox to approximate the neural
word-piece model on the word-piece WFA topology. We then convert the resulting
word-piece WFA LM to the equivalent n-gram LM. The remainder of this section
outlines efficient algorithms for converting between word and word-piece WFA
models.

A natural way to represent the transduction from word-piece sequences
to word sequences is with a finite-state transducer. Given the properties of our
word-piece representation, that transducer can be made sequential (i.e., input
deterministic).

A {\em sequential weighted finite-state transducer} (WFST) is a deterministic
WFA where each transition has an output label in addition to its (input) label
and weight. We will denote by $o_q[x]$ the output label of the transition at
state $q$ with input label $x$, $o_q[x] \in \Delta \cup \{\epsilon\}$, where
$\Delta$ denotes the output alphabet of the transducer and $\epsilon$ the
empty string/sequence.

Let $M$ be the minimal sequential (unweighted) finite-state transducer (FST)
lexicon from word-piece sequences in $\Sigma^*$ to word sequences in $\Delta^*$,
where $\Sigma$ denotes our word-piece inventory, $\Delta$ denotes our
vocabulary, and $*$ is Kleene closure.
\begin{algorithm*}[t]
  \caption{Approximating a Neural Model as an N-Gram with a Supplemental
           Topology.}
  \label{alg:workflow}
  \begin{varwidth}[t]{0.49\textwidth}
       \begin{algorithmic}
            \State Train $R_W^u$, $R_P^u$ with \texttt{FederatedAveraging}\footnote{
$T$ denotes an unweighted topology and $A$ denotes the weighted n-gram model.
Superscript $u$ represents uncased models.}
            \State Train $A_W$ from supplemental corpus $\mathbb{C}$
            \State $A_{W_e}$,$A_{W_i}$,$A_{W_m}$,$A_{W_r}$ $\gets$ $\text{Gen}(R_W^u,A_W,\o, \text{\NNFST}_\text{W}$)
            \State $A_{P_e}$,$A_{P_i}$,$A_{P_m}$,$A_{P_r}$ $\gets$ $\text{Gen}(R_P^u,A_W,A_{W_i}, \text{\NNFST}_\text{P}$)
            \State
            \Function{Gen}{$R^u$, $A_W$, $A_{W_i}$, function \NNFST}{}
              \State $A_e$ $\gets$ \NNFST($R^u$, $A_W$)
              \If {\NNFST==$\text{\NNFST}_\text{W}$}
                \State $A_i$ $\gets$ \NNFST($R^u$, $A_W$, self\_infer=true)
              \Else
                \State $A_i$ $\gets$ \NNFST($R^u$, $A_{W_i}$)
              \EndIf
              \State $A_m$ $\gets$ Interpolate($A_e$, $A_i$)
              \algstore{myalg}
        \end{algorithmic}
    \end{varwidth}\quad              
    \begin{varwidth}[t]{0.49\textwidth}
        \begin{algorithmic}
            \algrestore{myalg}
              \State $A_r$ $\gets$ \NNFST($R^u$, $A_m$)
              \State return $A_e$, $A_i$, $A_m$, $A_r$
            \EndFunction
            \Function{$\text{\NNFST}_\text{W}$}{$R_W^u$, $A_W$, self\_infer=false}{}
              \If {self\_infer}
                \State return \texttt{\Uapprox}($R_W^u$, $\o$, $A_W$)
              \Else
                \State return \texttt{\Uapprox}($R_W^u$, $A_W$, $A_W$)
              \EndIf
            \EndFunction
            \Function{$\text{\NNFST}_\text{P}$}{$R_P^u$, $A_{W}$}{}
              \State $T_W^u$ $\gets$ ConvertToLowercaseTopology($A_W$)
              \State $T_P^u$ $\gets$ ConvertToWordPieceTopology($T_W^u$)
              \State $A_P^u$ $\gets$ \texttt{\Wapprox}($R_P^u$, $T_P^u$)
              \State $A_W^u$ $\gets$ ConvertToWordTopology($A_P^u$)
              \State return \texttt{\Uapprox}($A_W^u$, $A_W$, $A_W$)
            \EndFunction
        \end{algorithmic}
    \end{varwidth}
\end{algorithm*}
A word-piece topology $B$ equivalent to the word topology $A$ can
be obtained by composing the word-piece-to-word transducer $M$ with $A$:
\begin{equation*}
B = M \circ A.
\end{equation*}
Since $A$ has backoff transitions, the generic composition algorithm
of~\cite{allauzen11} is used with a custom composition filter that ensures the
result, $B$, is deterministic with a well-formed backoff structure, and hence is
suitable for the counting step of \Wapprox. We give an explicit description of
the construction of $B$, from which readers familiar with~\citet{allauzen11} can
infer the form of the custom composition filter.

The states in $B$ are pairs $(q_1, q_2)$, with $q_1 \in Q_M$ and $q_2$
in $Q_A$, initial state $i_B = (i_M, i_A)$, and final state $f_B = (f_M, f_A)$.
Given a state $(q_1, q_2) \in Q_B$, the outgoing transitions and their
destination states are defined as follows. If $x \in L[q_1]$, then an
$x$-labeled transition is created if one of two conditions holds:
\begin{enumerate}[itemsep=0mm]
\item
if $o_{q_1}[x] \in L[q_2]$, then
\begin{align*}
d_{(q_1,q_2)}[x] &= (d_{q_1}[x], d_{q_2}[o_{q_1}[x]]) \text{ and }\\
o_{(q_1,q_2)}[x] &= o_{q_1}[x];
\end{align*}
\item
if $o_{q_1}[x] = \epsilon$ and $R[d_{q_1}[x]] \cap L[q_2] \not= \emptyset$,
then
\begin{align*}
d_{(q_1,q_2)}[x] &= (d_{q_1}[x], d_{q_2}[o_{q_1}[x]]) \text{ and }\\
o_{(q_1,q_2)}[x] &= \epsilon
\end{align*}
\end{enumerate}
where $R[q]$ denotes the set of output non-$\epsilon$ labels that can be emitted
after following an output-$\epsilon$ path from $q$. Finally if
$\varphi \in L[q_1]$, a backoff transition is created:
\begin{equation*}
d_{(q_1,q_2)}[\varphi] = (q_1, d_{q_2}[\varphi]) \text{ and }
o_{q_1,q_2}[\varphi] = \epsilon.
\end{equation*}
The counting step of \Wapprox is applied to $B$, and transfers the
computed counts from $B$ to $A$ by relying on the following key
property of $M$. For every word $y$ in $\Delta$, there exists a unique
state $q_y \in Q_M$ and unique word-piece $x_y$ in $\Sigma$ such that
$o_{q_y}[x_y] = y$. This allows us to transfer the counts from $B$ to $A$ as
follows:
\begin{equation*}
  w_q[y] = w_{(q_y, q)}[x_y]
\end{equation*}
The KL minimization step of \Wapprox to $A$ is applied subsequently.

As an alternative, the unweighted word automaton $A$ could be used to perform the
counting step directly. Each sample $\bar{x}(j)$ could be mapped to a
corresponding word sequence $\bar{y}(j)$, mapping out-of-vocabulary word-piece
sequences to an unknown token. However, the counting steps would have become
much more computationally expensive, since $p_{\text{nn}}(y|\bar{y}^i(j))$ would
have to be evaluated for all $i$, $j$ and for all words $y$ in the vocabulary,
where $p_{\text{nn}}$ is now a word-piece RNN.

\section{Experiments}
\label{sec:experiments}
\subsection{Neural language model}
\label{sec:nlm}
LSTM models~\cite{hochreiter1997long} have been successfully used in a variety
of sequence processing tasks. LSTM models usually have a large number of
parameters and are not suitable for on-device learning. In this work, we use
various techniques to reduce the memory footprint and to improve model performance.

We use a variant of LSTM with a Coupled Input and Forget Gate
(CIFG)~\cite{greff2017lstm} for the federated neural language model. CIFG
couples the forget and input decisions together, which reduces the number of
LSTM parameters by 25\%. We also use group-LSTM (GLSTM)~\cite{kuchaiev2017factorization} to reduce the
number of trainable variables of an LSTM matrix by the number of feature groups,
$k$.
We set $k=5$ in experiments.
\begin{table}[]
\centering
\begin{tabular}{c|cccccc}
Model                     & $N_l$ & $N_h$ & $N_e$ & $S_e$ & $S_{\text{total}}$     \\ \hline
$\text{W}_{\text{30K}}$   & 1     & 670   & 96    & 2.91M & 3.40M \\
$\text{P}_{\text{4K-S}}$  & 1     & 670   & 96    & 0.38M & 0.85M \\
$\text{P}_{\text{4K-L}}$  & 2     & 1080  & 140   & 0.56M & 2.70M \\
$\text{P}_{\text{4K-G}}$  & 2     & 1080  & 280   & 1.12M & 2.71M \\
$\text{P}_{\text{16K-S}}$ & 1     & 670   & 96    & 1.54M & 2.00M \\
$\text{P}_{\text{16K-L}}$ & 1     & 670   & 160   & 2.56M & 3.33M \\
$\text{P}_{\text{30K}}$   & 1     & 670   & 96    & 2.91M & 3.40M
\end{tabular}
\caption{Parameters for neural language models. W and P refer to word and
         word-piece models, respectively. $N_l$, $N_h$, $N_e$, $S_e$ and
         $S_{total}$ refer to the number of LSTM layers, the number of hidden
         states in LSTM, the embedding dimension size, the number of parameters
         in the embedding layer and in total, respectively. The suffixes ``S''
         and ``L'' indicate small and large models. ``G'' represents GLSTM. The
         suffixes 4K, 16K and 30K represent the vocabulary sizes.
         }
\label{table:modelarchitecture}
\end{table}
Table~\ref{table:modelarchitecture} lists the parameter settings of
the word (W) and word-piece (P) models used in this study. Due to the
memory limitations of on-device training, all models use fewer than
$3.5M$ parameters. For each vocabulary size, we first start with a
base architecture consisting of one LSTM layer, a $96$-dimensional
embedding, and $670$ hidden state units. We then attempt to increase
the representational power of the LSTM cell by increasing the number
of hidden units and using multi-layer LSTM
cells~\cite{sutskever2014sequence}. Residual
LSTM~\cite{kim2017residual} and layer
normalization~\cite{lei2016layer} are used throughout experiments, as
these techniques were observed to improve convergence.  To avoid the
restriction that $N_h=N_e$ in the output, we apply a projection step
at the output gate of the LSTM~\cite{sak2014long}. This step reduces
the dimension of the LSTM hidden state from $N_h$ to $N_e$.  We also
share the embedding matrix between the input embedding and output
softmax layer, which reduces the memory requirement by $\left
| \mathcal{V} \right | \times N_e$. We note that other recurrent
neural models such as \emph{gated recurrent
units}~\cite{chung2014empirical} can also be used instead of CIFG
LSTMs.

\begin{figure*}
  \centering
    \includegraphics[width=0.46\textwidth]{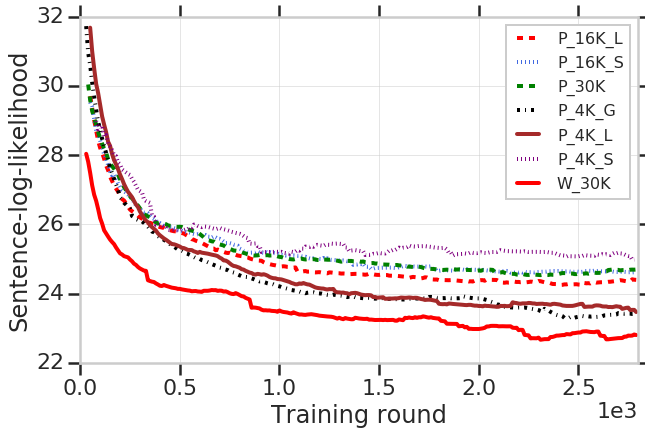}
  \centering
    \includegraphics[width=0.46\textwidth]{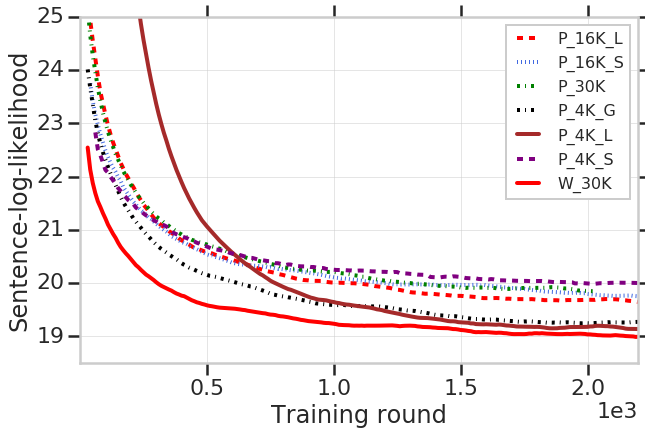}
  \caption{Sentence log likelihood excluding OOV token for en\_US (left) and pt\_BR (right).}
  \label{fig:sll}
\end{figure*}
The federated RNN LMs are trained on two language settings of
the virtual keyboard: American English (en\_US) and Brazilian
Portuguese (pt\_BR).
Following~\citet{mcmahan2016communication}, 500 reporting
clients are used to compute the gradient updates for each round. A
server-side learning rate of 1.0, a client-side learning rate of 0.5,
and Nesterov momentum of 0.9 are used. Both the word and word-piece models are
trained over the same time range and with the same hyperparameters.
Prior to federated training of the RNN LM, the word-piece
inventory is constructed
from the unigram distribution collected via the federated approach introduced in
Section~\ref{sec:unigram}.

A common evaluation metric for both word and word-piece models is
desirable during federated training. Such a metric can be used to
monitor the training status and select models to be used for the
\Uapprox algorithm. Neither cross-entropy nor accuracy serves this
need due to the mismatch in vocabularies used. Word-level accuracy is
hard to compute for the word-piece model, since it requires hundreds
of inference calls to traverse all combinations of a word from the
word-piece vocabulary.  In this study, we apply sentence log
likelihood (SLL) in the evaluation. Given a sentence $\bar{x}^m=\{x_1,
x_2, \ldots, x_m\}$ composed of $m$ units (either words or
word-pieces), SLL is evaluated as
$\sum_{i=1}^{m}\log(p_{nn}(x_i|\bar{x}^{i-1}))$. One issue that arises
is the handling of out-of-vocabulary (OOV) words. The OOV probability
of the word model is about $8\%$. The comparable probability of an OOV
word (according to $\mathcal{V}$) for word-piece models is the product
of the corresponding word-piece conditional probabilities, which is
much smaller than $8\%$. To mitigate this issue, we define SLL
excluding OOV as:
\[
\text{SLL}^e=\sum^m_{i : x_i\neq\text{OOV}}\log(p_{nn}(x_i|\bar{x}^{i-1})),
\]
where the OOV in the equation includes word-pieces that are components
of OOV words. In the following, $\text{SLL}^e$ is used as model
selection metric.

\subsection{Approximated n-gram model}

Algorithm~\ref{alg:workflow} illustrates the workflow we use to
generate different n-gram models for evaluation. Recall that \Uapprox
takes a RNN LM, an n-gram topology, and a reweighting FST for
capitalization normalization. The n-gram topology is empty under
self-inference mode. \citet{suresh2018approx} showed that
inferring topology from the RNN LM does not perform as well as
using the true n-gram topology obtained from the training corpus. Hence, we
supplement the neural-inferred topology with the topology
obtained by a large external large corpus denoted by $A_W$.
 We use
\Uapprox on four topologies and compare the resulting models: an n-gram model
obtained from an external corpus's topology $A_e$, an n-gram model obtained
from a neural inferred topology $A_i$, an n-gram model obtained by
interpolating (merging) the two models above $A_m$, and an n-gram model obtained
by approximating on the interpolated topology $A_r$. We repeat this
experiment for both word and word-piece RNN LMs and use subscripts $W$
and $P$, respectively.
We evaluate all eight produced n-gram models directly on the
traffic of a production virtual keyboard, where prediction accuracy is
evaluated over user-typed words.

\subsection{Results}
\begin{table}[]
\centering
\begin{tabular}{c|c|c}
Model                     & en\_US                         & pt\_BR       \\ \hline
Baseline                  & $10.03\%$                       & $8.55\%$       \\ \hline
$A_{W_e}$               & ${10.52\pm0.03\%}$      & $9.66\pm0.02\%$       \\
$A_{W_i}$               & $10.47\pm0.02\%$               & $9.67\pm0.02\%$      \\
$A_{W_m}$               & $10.27\pm0.03\%$               & ${9.40\pm0.02\%}$      \\
$A_{W_r}$               & $10.49\pm0.03\%$               & $9.65\pm0.02\%$      \\
\end{tabular}
\caption{Result of top-1 prediction accuracy on the live traffic of the virtual
         keyboard for en\_US and pt\_BR populations. Quoted 95\% confidence
         intervals for federated models are derived using the jackknife method.
         }
\label{table:result-ab}
\end{table}
\begin{table}[]
\centering
\begin{tabular}{c|c}
Model                     & top-1                   \\ \hline
Baseline                  & $10.03\%$              \\ \hline
$A_{P_e}$               & ${10.49\pm0.03\%}$      \\
$A_{P_i}$               & $10.46\pm0.03\%$               \\
$A_{P_m}$               & $10.48\pm0.04\%$             \\
$A_{P_r}$               & ${10.53\pm0.03\%}$              \\
\end{tabular}
\caption{Result of top-1 prediction accuracy on the live traffic of the virtual
         keyboard for en\_US derived using word-piece models.}
\label{table:result-ab-wp}
\end{table}
Figure~\ref{fig:sll} shows the $\text{SLL}^e$ metric for all the
experiments listed in Table~\ref{table:modelarchitecture}. In
general, larger models generate better results than smaller
baseline models. For the baseline architectures with same RNN
size, having a larger vocabulary leads to some gains. For the larger
architectures that have similar total numbers of parameters, 4K
word-piece models are shown to be superior to 16K and 30K. For 4K
word-piece models, GLSTM is in general on-par with its
$\text{P}_{\text{4K-L}}$ counterpart. The word model is better than
all the word-piece models in both languages in $\text{SLL}^e$. We were surprised
by this result, and hypothesize that it is due to the $\text{SLL}^e$ metric
discounting word-piece models' ability to model the semantics of OOV words. The
solid lines are the best models we pick for A/B experiment evaluation for the
virtual keyboard ($\text{P}_{\text{4K-L}}$ and $\text{W}_{\text{30K}}$).

Table~\ref{table:result-ab} shows the A/B evaluation result on both en\_US
and pt\_BR populations. The baseline model is an n-gram model
trained directly from centralized logs. All of the federated trained models
perform better than the baseline model. We repeated the A/B evaluation with word-piece models on
en\_US and the results are in Table~\ref{table:result-ab-wp}.
The performance of word-piece models is similar to that of word models.
Among the federated models for en\_US, $A_{P_r}$ has the best result. This meets our expectation that the
supplemental corpus helps improve the performance of the topology inferred from
the RNN LM.

\section{Conclusion}
\label{sec:conclusion}
We have proposed methods to train production-quality n-gram language models
using federated learning, which allows training models without user-typed text ever leaving devices.
The proposed methods are shown to perform better than traditional server-based algorithms in A/B experiments on real users of a virtual keyboard.

\section*{Acknowledgments}
The authors would like to thank colleagues in Google Research
for providing the federated learning framework and for
many helpful discussions.

\bibliography{conll-2019}
\bibliographystyle{acl_natbib}

\end{document}